\RequirePackage{amsmath}
\documentclass[runningheads,a4paper]{llncs}

\usepackage[utf8]{inputenc} 
\usepackage{graphicx}
\usepackage{xcolor}
\usepackage[color=blue!20, linecolor=blue!80!black]{todonotes}
\usepackage[english]{babel}
\usepackage[utf8]{inputenc}
\DeclareUnicodeCharacter{00A0}{ }
\usepackage{hyperref}
\usepackage{xspace}
\usepackage{amsmath} 
\usepackage[capitalise,nameinlink]{cleveref}
\usepackage[binary,squaren]{SIunits} 
\usepackage{amssymb}
\usepackage{listings}
\usepackage{dsfont}
\usepackage{multicol}
\usepackage{cite}
\usepackage{float}
\usepackage{booktabs}
\usepackage{cleveref}
\usepackage{pgfplots}
\usepackage{caption}
\usepackage{subcaption}
\usepackage{soul}

\usepackage{booktabs} 
\usepackage{xspace} 
\usepackage{lstautogobble}  

\usepackage{underscore}

\usetikzlibrary{shapes, arrows, positioning, calc}
\newcommand*\streamlab{\textsc{StreamLAB}\xspace}
\newcommand*\rtlola{\textsc{RTLola}\xspace}

\newcommand*\fpga{\text{FPGA}\xspace}
\newcommand*\vhdl{\text{VHDL}\xspace}
\newcommand*\asic{\text{ASIC}\xspace}

\newcommand*\todocite[1]{[?]\todo{Cite #1}}
\definecolor{CommentColor}{RGB}{42,0.0,255} 

\makeatletter
\newcommand*{\etal}{%
    \@ifnextchar{.}%
        {et~al}%
        {et~al.\@\xspace}%
}
\newcommand*{\ie}{i.e.\@\xspace}
\newcommand*{\eg}{e.g.\@\xspace}
\newcommand*{\wrt}{w.r.t.\@\xspace}
\makeatother

\definecolor{bluekeywords}{rgb}{0.13, 0.13, 1}
\definecolor{greentypes}{rgb}{0, 0.5, 0}
\definecolor{redstrings}{RGB}{171, 114, 2}
\definecolor{graynumbers}{rgb}{0.5, 0.5, 0.5}
\definecolor{goldcomments}{rgb}{0.6, 0.4, 0.08}
\definecolor{monitorblue}{RGB}{18, 163, 38}

\lstdefinelanguage{Lola}{
  keywords=[0]{input, output, trigger, import},
  keywordstyle=[0]\bfseries\color{bluekeywords},
  keywords=[1]{if, then, else, aggregate, defaults, offset},
  keywords=[2]{Int8, Int16, Int32, Int64, UInt8, UInt16, UInt32, UInt64, Bool, Float16, Float32, Float64, @1Hz, @5Hz, @10Hz, @100mHz, @1kHz},
  keywordstyle=[2]\color{greentypes},
    sensitive=false,
    comment=[l]{//},
    morecomment=[s]{/*}{*/},
    morestring=[b]',
    morestring=[b]"
}

\begin{document}
\hyphenation{Time-stamp}
\hyphenation{Au-to-no-mous}
\hyphenation{RTLola}

\title{RTLola Cleared for Take-Off: \\Monitoring Autonomous Aircraft\thanks{This work was partially supported by the German Research Foundation (DFG) as part of the Collaborative Research Center Foundations of Perspicuous Software Systems (TRR 248, 389792660), and by the European Research Council (ERC) Grant OSARES (No. 683300).}}
\titlerunning{RTLola Cleared for Take-Off}

\author{Jan Baumeister\inst{1}\orcidID{0000-0002-8891-7483}\and
Bernd Finkbeiner\inst{1}\orcidID{0000-0002-4280-8441} \and
Sebastian Schirmer\inst{2} \and \\
Maximilian Schwenger\inst{1}\orcidID{0000-0002-2091-7575} \and
Christoph Torens\inst{2}
}

\authorrunning{Baumeister et al.}
\titlerunning{RTLola Cleared for Take-Off}
\institute{%
  Saarland University, Department of Computer Science, \\
  66123 Saarbr\"ucken, Germany \\
  \email{\{jbaumeister, finkbeiner, schwenger\}@react.uni-saarland.de} \and
  German Aerospace Center (DLR), \\
  38108 Braunschweig, Germany \\
  \email{\{christoph.torens, sebastian.schirmer\}@dlr.de}%
}

\maketitle

\begin{abstract}
  The autonomous control of unmanned aircraft is a highly
  safety-critical domain with great economic potential in a wide range
  of application areas, including logistics, agriculture, civil engineering, and
  disaster recovery.  We report on the development of a dynamic
  monitoring framework for the DLR ARTIS (Autonomous Rotorcraft Testbed
  for Intelligent Systems) family of unmanned aircraft based on the
  formal specification language RTLola.  RTLola is a stream-based
  specification language for real-time properties. An RTLola
  specification of hazardous situations and system failures is
  statically analyzed in terms of consistency and resource usage and
  then automatically translated into an FPGA-based monitor. Our
  approach leads to highly efficient, parallelized monitors with formal
  guarantees on the noninterference of the monitor with the normal
  operation of the autonomous system.

\end{abstract}

\section{Introduction}
\label{sec:intro}

An unmanned aerial vehicle, commonly known as a drone, is an aircraft
without a human pilot on board. While usually connected via radio
transmissions to a base station on the ground, such aircraft are
increasingly equipped with decision-making capabilities that allow
them to autonomously carry out complex missions in applications such as
transport, mapping and surveillance, or crop and irrigation
monitoring.  Despite the obvious safety-criticality of such systems,
it is impossible to foresee all situations an autonomous aircraft
might encounter and thus make a safety case purely by analyzing all
of the potential behaviors in advance.  A critical part of the safety
engineering of a drone is therefore to carefully monitor the actual
behavior during the flight, so that the health status of the system can
be assessed and mitigation procedures (such as a return to the base
station or an emergency landing) can be initiated when needed.

In this paper, we report on the development of a dynamic monitoring
framework for the DLR ARTIS (Autonomous Rotorcraft Testbed for
Intelligent Systems) family of aircraft based on the formal
specification language \rtlola.  The development of a monitoring
framework for an autonomous aircraft differs significantly from a
monitoring framework in a more standard setting, such as network
monitoring. A key consideration is that while the specification
language needs to be highly \emph{expressive}, the monitor must
operate within strictly limited resources, and the monitor itself needs
to be highly \emph{reliable}: any interference with the normal
operation of the aircraft could have fatal consequences.

A high level of expressiveness is necessary because the assessment of
the health status requires complex analyses, including a
cross-validation of different sensor modules such as the agreement
between the GPS module and the accelerometer. This is necessary in
order to discover a deterioration of a sensor module. At the same
time, the expressiveness and the precision of the monitor must be
balanced against the available computing resources.  The reliability
requirement goes beyond pure correctness and robustness of the
execution.  Most importantly, reliability requires that the peak
resource consumption of the monitor in terms of energy, time, and
space needs to be known ahead of time.  This means that it must be
possible to compute these resource requirements statically based on an
analysis of the specification.  The
determination whether the drone is equipped with sufficient hardware
can then be made before the flight, and the occurrence of dynamic failures
such as running out of memory or sudden drops in voltage can be ruled
out. Finally, the collection of the data from the on-board
architecture is a non-trivial problem: While the monitor needs access
to almost the complete system state, the data needs to be retrieved
non-intrusively such that it does not interfere with the normal system
operation.

Our monitoring approach is based on the formal stream specification language \rtlola~\cite{DBLP:conf/cav/FaymonvilleFSSS19}. In an \rtlola specification, input streams that collect data from sensors, networks, etc.,
are filtered and combined into output streams that contain data
aggregated from multiple sources and over multiple points in time such
as over sliding windows of some real-time length. Trigger
conditions over these output streams then identify critical
situations. An \rtlola specification is translated into a monitor defined in a  hardware description language and subsequently realized on an \fpga.
Before deployment, the specification is checked for consistency and the minimal requirements on the \fpga are computed.
The hardware monitor is then placed in a central position where as much sensor data as possible can be collected; during the execution, it then extracts the relevant information.
In addition to requiring no physical changes to the system architecture, this integration incurs no further traffic on the bus.

Our experience has been extremely positive: Our approach leads to highly efficient, parallelized monitors with formal guarantees on the non-interference of the monitor with the normal operation of the autonomous system.
The monitor is able to detect violations to complex specifications without intruding into the system execution, and operates within narrow resource constraints. \rtlola is cleared for take-off.

\newcommand\previousversion[1]{}
\previousversion{
Cyber-physical systems (CPS) carry out increasingly safety-critical tasks, such as controlling medical CPS, power plants, and autonomous vehicles. 
The complexity of these systems often exceeds the scalability of static verification techniques.
This renders dynamic verification methods such as runtime monitoring inevitable, as can be seen in the large body of recent works in this field\todocite{many recent papers}.
In these approaches, rather than verifying the correctness of all possible executions of a system \wrt a formal specification, a single execution --- the current execution of the system --- is inspected during runtime.
For this, sensors capture the state of the system, which allows for assessing the health of the system and gaining an insight into its behavior.
Upon detection of undesired behavior, an alarm is raised upon which mitigation procedures can be initiated.

There are multiple requirements on a runtime verification framework: 
Not only does the input language of the monitor need to be sufficiently \emph{expressive}, the monitor itself needs to be \emph{reliable} as well.
A high level of expressiveness is necessary because the assessment of a CPS's health requires complex analyses.
They need to answer questions like ``To which degree do measurements of the GPS module agree with readings of the accelerometer?'', which allows for a cross-validation of different sensor modules. 
This way, the deterioration of a sensor module can be discovered and subsequently replaced by a spare module --- a technique widely used in systems with a high level of redundancy such as aircraft.
However, a high level of expressiveness in the specification language comes at the cost of high resource consumption of the monitor. 
However, most embedded devices do not have exuberant amounts of spare resources, nor can the main functionality be significantly reduced to create them. 
Thus, developers of runtime verification techniques need to find the sweet spot between adequate expressiveness and resource consumption, \eg by forsaking generality for specific, highly relevant domain-specific language features.

The reliability of the monitor is independent of the expressiveness of the input language and goes beyond pure correctness and robustness of the execution.    
Most notably, this additionally entails that its peak resource consumption in terms of energy, time, and space needs to be available statically. 
This way, engineers can evaluate whether their hardware can cope with the requirements of the monitor before deployment.
This increases the confidence in the monitor by ruling out dynamic failures such as running out of memory or sudden drops in voltage.\todo{We can elaborate here or earlier that static criteria are far superior to dynamic ones which superficially contradicts the idea that RV $>$ MC.  However, we can also just leave it out.}

Lastly, the integration of the monitor is a delicate endeavor:
While the monitor requires as much information about the system state as possible, it may not require a complete overhaul of the system architecture.
Moreover, in many systems, sensor data needs to be retrieved non-intrusively such that it does not mingle with the regular execution of the system. 

In this paper we report on the integration of the \streamlab monitoring framework into an existing autonomous drone.  
The monitor is based on an input specification in the \rtlola language, which is translated into a hardware description language and realized on an \fpga.
Before deployment, the specification is analyzed for inner consistency and the minimal requirements on the \fpga are computed.
The hardware monitor is then placed in a central position where as much sensor data as possible can be collected; during the execution, it then extracts the relevant information.
In addition to requiring no physical changes to the system architecture, this integration approach incurs no further traffic on the bus.
We found that the monitor is able to detect violations to complex specifications without intruding into the system execution, nor requiring unreasonable resources.
}

\subsection{Related Work}
\label{sec:rw}

Stream-based monitoring approaches focus on an expressive specification language while handling non-binary data.
Its roots lie in synchronous, declarative stream processing languages like Lustre~\cite{lustre} and Lola~\cite{oldlola}.
The \emph{Copilot} framework \cite{Pike2010} features a declarative data-flow language from which constant space and constant time C monitors are generated; these guarantees enable usage on an embedded device.
Rather than focusing on data-flow, the family of Lola-languages puts an emphasis on statistical measures and has successfully been used to monitor synchronous, discrete time properties of autonomous aircraft~\cite{uav1,uav2}.
In contrast to that, \rtlola~\cite{rtlolaarxiv,Schwenger/19} supports real-time capabilities and efficient aggregation of data occurring with arbitrary frequency, while forgoing parametrization for efficiency~\cite{DBLP:conf/cav/FaymonvilleFSSS19}. 
\rtlola can also be compiled to \vhdl and subsequently realized on an \fpga~\cite{fpgalola}.

Apart from stream-based monitoring, there is a rich body of monitoring based on real-time temporal logics~\cite{xxxA,Koymans1990,Raskin1997,DBLP:conf/lics/HarelLP90,Donze:2010:RST:1885174.1885183,6313045} such as Signal Temporal Logic (STL)~\cite{DBLP:journals/sttt/MalerN13}.
Such languages are a concise way to describe temporal behaviors with the shortcoming that they are usually limited to qualitative statements, \ie boolean verdicts.
This limitation was addressed for STL~\cite{DBLP:conf/formats/DonzeM10} by introducing a quantitative semantics indicating the robustness of a satisfaction.
To specify continuous signal patterns, specification languages based on regular expressions can be beneficial, e.g.~Signal Regular Expressions (SRE)~\cite{DBLP:conf/formats/BakhirkinFMU17}. 
The R2U2 tool~\cite{DBLP:journals/fmsd/MoosbruggerRS17} stands out in particular as it successfully brought a logic closely related to STL onto unmanned aerial systems as an external hardware implementation.
%

\section{Setup}\label{sec:setup}
The Autonomous Rotorcraft Testbed for Intelligent Systems (ARTIS) is a platform used by the Institute of Flight Systems of the German Aerospace Center (DLR) to conduct research on autonomous flight.
It consists of a set of unmanned helicopters and fixed-wing aircraft of different sizes which can be used to develop new techniques and evaluate them under real-world conditions.

The case study presented in this paper revolves around the superARTIS, a large helicopter with a maximum payload of $85\kilo\gram$, depicted in \cref{fig:superartis}. 
The high payload capabilities allow the aircraft to carry multiple sensor systems, computational resources, and data links.
This extensive range of avionic equipment plays an important role in improving the situational awareness of the aircraft~\cite{ammann2017} during the flight.
It facilitates safe autonomous research missions which include flying in urban or maritime areas, alone or with other aircraft.   
Before an actual flight test, software- and hardware-in-the-loop simulations, as well as  real-time logfile replays strengthen confidence in the developed technology.

\subsection{Mission}\label{sec:paths}
One field of application for unmanned aerial vehicles (UAVs) is reconnaissance missions.
In such missions, the aircraft is expected to operate within a fixed area in which it can cause no harm. 
The polygonal boundary of this area is called a geo-fence.
As soon as the vehicle passes the geo-fence, mitigation procedures need to be initiated to ensure that the aircraft does not stray further away from the safe area.

\begin{figure}[t]
\begin{minipage}{\linewidth}
   	\begin{minipage}{0.47\linewidth}
   	  \centering
    	\begin{figure}[H]
    	\centering
          \includegraphics[width=.8\linewidth]{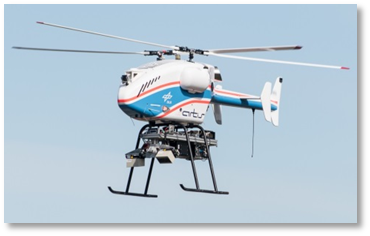}
          \caption{DLR's autonomous superARTIS equipped with optical navigation.}
          \label{fig:superartis}       
       	\end{figure}
 	\end{minipage}
 	\hfill
 	\begin{minipage}{0.47\linewidth}
 	   	\centering
       	\begin{figure}[H]
       	\centering 
          \includegraphics[width=.8\linewidth]{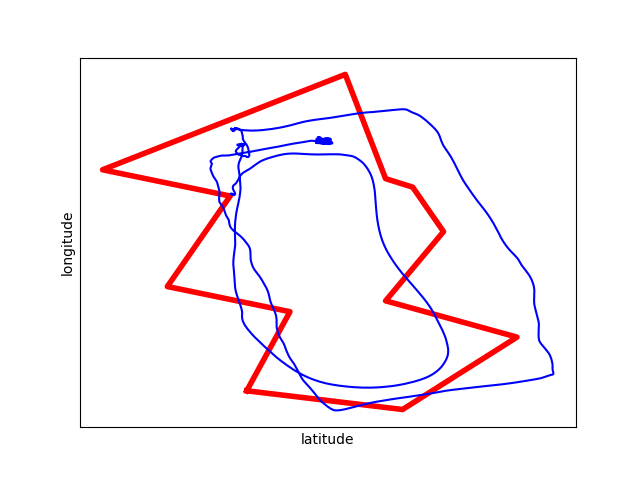}
          \caption{Reconnaissance mission for a UAV. The thin blue line represents its trajectory, the thick red line a geo-fence.}
          \label{fig:mission}
      	\end{figure}
	\end{minipage}
\end{minipage}
\end{figure}

The case study presented in this paper features a reconnaissance mission. 
\cref{fig:mission} shows the flight path (blue line) within a geo-fence (red line). 
Evidently, the aircraft violates the fence several times temporarily.
A reason for this can be flawed position estimation:  An aircraft estimates its position based on several factors such as landmarks detected optically or GPS sensor readings.
In the latter case, GPS satellites send position and time information to earth. 
The GPS module uses this data to compute the aircraft's absolute position with trilateration. 
However, signal reflection or a low number of GPS satellites in range can result in imprecisions in the position approximation.
If the aircraft is continuously exposed to imprecise position updates, the error adds up and results in a strong deviation from the expected flight path.

The impact of this effect can be seen in \cref{fig:acceleration}.
It shows the velocity of a ground-borne aircraft in an enclosed backyard according to its GPS module.\footnote{GPS modules only provide absolute position information; the first derivative thereof, however, is the velocity.}
During the reported period of time, the aircraft was pushed across the backyard by hand.
While the expected graph is a smooth curve, the actual measurements show an erratic curve with errors of up to $\pm 1.5\meter\reciprocal\second$, 
which can be mainly attributed to signals being reflected on the enclosure.
The strictly positive trend of the horizontal velocity can explain strong deviations from the desired flight path seen in \cref{fig:acceleration}.

A counter-measure to these imprecisions is the cross-validation of several redundant sensors. 
As an example, rather than just relying on the velocity reported by a GPS module, its measured velocity can be compared to the integrated output of an accelerometer. 
When the values deviate strongly, the values can be classified as less reliable than when both sensors agree.

\subsection{Non-Intrusive Instrumentation}\label{sec:logging}

When integrating the monitor into an existing system, the system architecture usually cannot be altered drastically.
Moreover, the monitor should not interfere with the regular execution of the system, \eg by requiring the controller to send explicit messages to it.
Such a requirement could offset the timing behavior and thus have a negative impact on the overall performance of the system.

\begin{figure}[t]
\begin{minipage}{\linewidth}
   	\begin{minipage}[b]{0.47\linewidth}
    	\begin{figure}[H]
          \includegraphics[width=\linewidth]{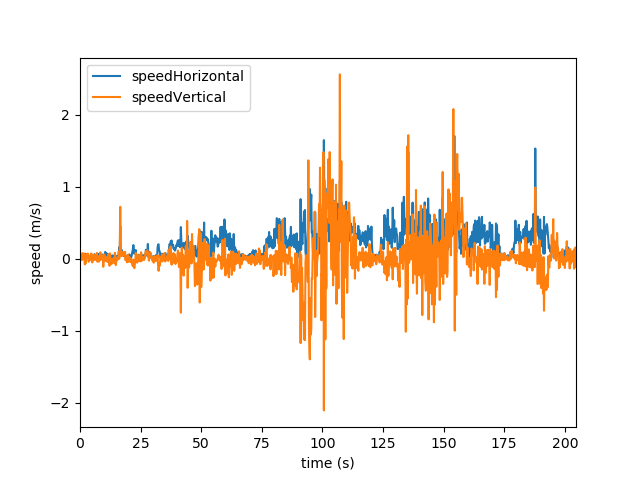}
          \caption{Line plot of the horizontal and vertical speed calculated by a GPS receiver.}
          \label{fig:acceleration}
       	\end{figure}
 	\end{minipage}
	\hfill
 	\begin{minipage}[b]{0.47\linewidth}
       	\begin{figure}[H]
         	\centering
          	\definecolor{mydarkblue}{rgb}{0.412,0.475,0.957}
\tikzstyle{state} = [circle, draw, text width=.5cm, text centered, minimum height=.5cm, minimum width=.5cm]
\tikzstyle{box} = [rectangle, draw, text centered, minimum height=1cm, minimum width=1cm, fill=white]
\tikzstyle{astate} = [circle, draw, text width=1.5cm, text centered, minimum height=2cm, minimum width=2cm]
\tikzstyle{smallcomponent} = [rectangle, draw, text width=5.5em, text centered, minimum height=3em, fill=white]
\tikzstyle{component} = [rectangle, draw, text width=8em, text centered, minimum height=3em, fill=white]
\tikzstyle{smcomponent} = [component, minimum width=3cm, minimum height=2.5cm]
\tikzstyle{transition} = [draw, -stealth']
\tikzstyle{line} = [draw]
\tikzstyle{signal} = [draw, -latex']
\tikzstyle{config} = [densely dotted]

\colorlet{eventcolor}{green!50!black}
\colorlet{periodiccolor}{blue!50!black}
\tikzstyle{event} = [draw=eventcolor, thin, fill=white,fill opacity=.3, pattern=north west lines, pattern color=eventcolor]
\tikzstyle{periodic} = [draw=periodiccolor, thin, fill opacity=.3, pattern=north east lines, pattern color=periodiccolor]
\tikzstyle{signalname} = [text width=10em, minimum height=2em]
\tikzstyle{nameright} = [signalname, align=left]
\tikzstyle{namecenter} = [minimum height=2em, align=center, rotate=65]
\tikzstyle{nameleft} = [signalname, align=right]

\begin{tikzpicture}[scale=0.5, every node/.style={transform shape}]
  
    \node [smallcomponent, text width=1.8cm] at (2.8,2.1)  (dots) {\textsc{\dots}};
    \node [smallcomponent] at (1.4,0.9)  (imu) {\textsc{IMU}};
    \node [smallcomponent] at (0,2.1) (gnss) {\textsc{GNSS}};
    \node [smallcomponent] at (-1.4,0.9) (lidar) {\textsc{Lidar}};
    \node [smallcomponent] at (-2.8,2.1) (camera) {\textsc{Camera}};
    \node [box] at (1.5,-1.7) (alg) {\textsc{PositionAlgorithm}};
    \node [box,minimum width=4cm, minimum height=3em,] at (0,-3.4) (logger) {\textsc{Logger}};
    \node [box,draw=monitorblue, minimum height=3em,thick] at (-1.5,-5.1) (mon) {\textcolor{monitorblue}{\textsc{Monitor}}};
    \node [box,minimum height=3em,] at (1.5,-5.1) (disk) {\textsc{HardDisk}};

	\path [line,line width=0.5mm] (-3.5,0) -- (3.5,0);
	\path [signal] (dots) -- (2.8,0);
	\path [signal] (imu) -- (1.4,0);
	\path [signal] (gnss) -- (0,0);
	\path [signal] (lidar) -- (-1.4,0);
	\path [signal] (camera) -- (-2.8,0);
	\path [signal,line width=0.5mm] (0,0) -- (0,-0.5) -- (-1.5,-0.5) -- (-1.5,-2.9);
	\path [signal,line width=0.5mm] (0,-0.5) -- (1.5,-0.5) -- (alg);
	\path [signal,line width=0.5mm] (alg) -- (1.5,-2.9);
	\path [signal,line width=0.5mm] (1.5,-3.9) -- (disk);
	\path [signal,line width=0.5mm, color=monitorblue] (-1.5,-3.9) -- (mon);
    
  \end{tikzpicture}
  
%
    
            \caption{Overview of data flow in system architecture.}
            \label{fig:vehiclearchitecture}
      	\end{figure}
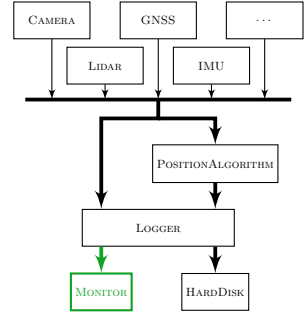
	\end{minipage}
	\vspace{1em}
\end{minipage}
\end{figure}

The issue can be circumvented by placing the monitor at a point where it can access all data necessary for the monitoring process non-intrusively.
In the case of the superARTIS, the logger interface provides such a place as it compiled the data of all position-related sensors as well as the output of the position estimation~\cite{ammann2017,ammann22017}.
\cref{fig:vehiclearchitecture} outlines the relevant data lines of the aircraft. 
Sensors were polled with fixed frequencies of up to $100\hertz$. 
The schematic shows that the logger explicitly sends data to the monitor.
This is not a strict requirement of the monitor as it could be connected to the data busses leading to the logger and passively read incoming data packets.
However, in the present setting, the logger did not run at full capacity. 
Thus sending information to the monitor came at no relevant cost while requiring few hardware changes to the bus layout.

In turn, the monitor provides feedback regarding violations of the specification. 
Here, we distinguish between different timing behaviors of triggers. 
The monitor evaluates event-based triggers whenever the system passes new events to the monitor and immediately replies with the results.
For periodic triggers, \ie, those annotated with an evaluation frequency, the evaluation is decoupled from the communication between monitor and system.
Thus, the monitor needs to wait until it receives another event until reporting the verdict.  
This incurs a short delay between detection and report.

\noindent


\subsection{StreamLAB}\label{sec:streamlab}
\streamlab\footnote{\url{stream-lab.eu}}\cite{DBLP:conf/cav/FaymonvilleFSSS19} is a monitoring framework revolving around the stream-based specification language \rtlola. 
It emphasizes on analyses conducted before deployment of the monitor. 
This increases the confidence in a successful execution by providing information to aid the specifier.
To this end, it detects inconsistencies in the specification such as type errors, \eg an lossy conversion of a floating point number to an integer, or timing errors, \eg accessing values that might not exist. 
Further, it provides two execution modes: an interpreter and an \fpga compilation.  
The interpreter allows the specifier to validate their specification. 
For this, it requires a \emph{trace}, \ie a series of data that is expected to occur during an execution of the system.
It then checks whether a trace complies with the specification and reports the points in time when specified bounds are violated.
After successfully validating the specification, it can be compiled into \vhdl code.
Yet again, the compiled code can be analyzed with respect to the space and power consumption.
This information allows for evaluating whether the available hardware suffices for running the \rtlola monitor.

An \rtlola specification consists of input and output streams, as well as trigger conditions.
\emph{Input} streams describe data the system produces asynchronously and provides to the monitor. 
\emph{Output} streams use this data to assess the health state of the system \eg by computing statistical information.
\emph{Trigger} conditions distinguish desired and undesired behavior. 
A violation of the condition issues an alarm to the system.

The following specification declares a floating point input stream \lstinline{height} representing sensor readings of an altimeter.
The output stream \lstinline{avg_height} computes the average value of the \lstinline{height} stream over two minutes.  
The aggregation is a sliding window computed once per second, as indicated with the \lstinline{@1Hz} annotation.\footnote{Details on how such a computation can cope with a statically-bounded amount of memory can be found in \cite{rtlolaarxiv,Schwenger/19}.} 
The stream \lstinline{$\delta$height} computes the difference between the average and the current height. 
A strong deviation of these values constitutes a suspicious jump in sensor readings, which might indicate a faulty sensor or an unexpected loss or gain in height.
In this case, the trigger in the specification issues a warning to the system, which can initiate mitigation measures.
\begin{lstlisting}
  input height: Float32
  output avg_height @1Hz := height.aggregate(over: 2min, using: avg)
  output $\delta$height := abs(avg_height.hold().defaults(to: height) - height)
  trigger $\delta$height > 50.0 "WARNING: Suspicious jump in height."
\end{lstlisting}
Note that this is just a brief introduction to \rtlola and the \streamlab framework.  
For more details, the authors refer to \cite{Schwenger/19,rtlolaarxiv,DBLP:conf/cav/FaymonvilleFSSS19,fpgalola}.

\subsection{FPGA as Monitoring Platform}\label{sec:fpga}

An \rtlola specification can be compiled into the hardware description language \vhdl and subsequently realized on an \fpga as proposed by Baumeister~\etal\cite{fpgalola}.
An \fpga as target platform for the monitor has several advantages in terms of improving the development process, reducing its cost, and increasing the overall confidence in the execution.

Since the \fpga is a separate module and thus decoupled from the control software, these components do not share processor time or memory.
This especially means that control and monitoring computations happen in parallel.
Further, the monitor itself parallelizes the computation of independent \rtlola output streams with almost no additional overhead. 
This significantly accelerates the monitoring process~\cite{fpgalola}.
The compiled \vhdl specification allows for extensive static analyses. 
Most notably, the results include whether the board is sufficiently large in terms of look-up tables and storage capabilities to host the monitor, and the power consumption when idle or at peak performance.
Lastly, an \fpga is the sweet spot between generality and specificity: it runs faster, is lighter, and consumes less energy than general purpose hardware while retaining a similar time-to-deployment.
The latter combined with a drastically lower cost renders the \fpga superior to application-specific integrated circuits (\asic) during development phase.
After that, when the specification is fixed, an \asic might be considered for its yet increased performance.

\subsection{RTLola Specifications}\label{sec:specs}
The entire specification for the mission is comprised of three sub-specifications.  
This section briefly outlines each of them and explains representative properties in \cref{fig:crossspecAbbrev}.
The complete specifications as well as a detailed description were presented in earlier work~\cite{schirmer2018,baumeister20}.

\begin{description}
  \item[Sensor Validation (\cref{appendix:validation})] 
    Sensors can produce incorrect values, \eg when too few GPS satellites are in range for an accurate trilateration or if the aircraft flies above the range of a radio altimeter. 
    A simple exemplary validation is to check whether the measured altitude is non-negative. 
    If such a check fails, the values are meaningless, so the system should not take them into account in its computations.
  \item[Geo-Fence (\cref{appendix:geofence})] 
    During the mission, the aircraft has permission to fly inside a zone delimited by a polygon, called a geo-fence. 
    The specification checks whether a face of the fence has been crossed, in which case  the aircraft needs to ensure that it does not stray further from the permitted zone.
  \item[Sensor Cross-Validation (\cref{appendix:cross})] 
    Sensor redundancy allows for validating a sensor reading by comparing it against readings of other sensors. 
    An agreement between the values raises the confidence in their correctness.  
    An example is the cross-validation of the GPS module against the accelerometer. 
    Integrating the readings of the latter twice yields an absolute position which can be compared against the GPS position.
\end{description}

\cref{fig:crossspecAbbrev} points out some representative sub-properties of the previously described specification in \rtlola, which are too long to discuss them in detail. It contains a validation of GPS readings as well as a cross-validation of the GPS module against the Inertial Measurement Unit (IMU).  
The specification declares three input streams, the $x$-position and number of GPS satellites in range from the GPS module, and the acceleration in $x$-direction according to the IMU.

The first trigger counts the number of updates received from the GPS module by counting how often the input stream \lstinline{gps_x} gets updated to validate the timing behavior of the module.

The output stream \lstinline{few_sat} computes the indicator function for \lstinline{num_sat < 9}, which indicates that the GPS module might report unreliable data due to few satellites in reach.
If this happens more than 12 times within five seconds, the next trigger issues a warning to indicate that the incoming GPS values might be inaccurate.
The last trigger checks whether the double integral of the IMU acceleration coincides with the GPS position up to a threshold of 0.5 meters.

\begin{figure}[t]
	\centering
	\input{specifications/crossValidationAbbrev}
	\caption{An \rtlola specification validating GPS sensor data and cross validating readings from the GPS module and IMU.}
	\label{fig:crossspecAbbrev}
	\vspace{-.5cm}
\end{figure}

\subsection{VHDL Synthesis}\label{sec:board}
The specifications mentioned above were compiled into \vhdl and realized on the Xilinx ZC702 Base Board\footnote{\url{https://www.xilinx.com/support/documentation/boards_and_kits/zc702_zvik/ug850-zc702-eval-bd.pdf}}.
The following table details the resource consumption of each sub-specification reported by the synthesis tool Vivado.
\noindent\begin{center}
\begin{tabular}{@{} *9r @{}}
  \toprule
  Spec       & FF & FF[$\%$]   & LUT  & LUT[$\%$]  & MUX & Idle [$\milli\watt$] & Peak [$\watt$] \\
  \midrule
  Geo-fence  & 2,853 & 3 & 26,181 & 71 & 4   & 149 & 1.871 \\
  Validation & 4,792 & 5 & 34,630 & 67 & 104 & 156 & 2.085 \\
  Cross      & 3,441 & 4 & 23,261 & 46 & 99  & 150 & 1.911 \\
  \bottomrule
\end{tabular}	
\end{center}
The number of flip-flops (FF) indicates the memory consumption in bits; neither specification requires more than $600\byte$ of memory.
The number of LUTs (Look-up Tables) is an indicator for the complexity of the logic. 
The sensor validation, despite being significantly longer than the cross-validation, requires the least amount of LUTs. 
The reason is that its computations are simple in comparison: Rather than computing sliding window aggregations or line intersections, it mainly consists of simple thresholding.
The number of multiplexers (MUX) reflects this as well: Since thresholding requires comparisons, which translate to multiplexers, the validation requires twice as many of them.
Lastly, the power consumption of the monitor is extremely low: When idle, neither specification requires more than $156\milli\watt$ and even under peak pressure, the power consumption does not exceed $2.1\watt$.  
For comparison, a Raspberry Pi needs between $1.1\watt$ (Model 2B) and $2.7\watt$ (Model 4B) when idle and roughly twice as much under peak pressure, i.e., $2.1\watt$ and $6.4\watt$, respectively.\footnote{Information collected from \url{https://www.pidramble.com/wiki/benchmarks/power-consumption} in January, 2020.}

\begin{figure}[t]
     \centering
     \begin{subfigure}[b]{0.47\textwidth}
         \centering
            \begin{tikzpicture}[scale=.5]
              \begin{axis}[
                xlabel=Number of Faces,
                ylabel=Look-Up Tables
              ]
                \addplot coordinates {
                  (0,2038)
                  (1,8441)
                  (2,7369)
                  (3,10304)
                  (4,8737)
                  (5,14387)
                  (6,17350)
                  (7,20387)
                  (8,23270)
                  (9,24110)
                  (10,27182)
                  (11,28199)
                  (12,33641)
                  (13,38611)
                  (14,34789)
                };
              \end{axis}
            \end{tikzpicture}
         \caption{Look-Up Tabels}
         \label{fig:resconsumptionluts}
     \end{subfigure}
     \hfill
     \begin{subfigure}[b]{0.47\textwidth}
         \centering
            \begin{tikzpicture}[scale=.50]
              \begin{axis}[
                xlabel=Number of Faces,
                ylabel=Worst Negative Slack Time
              ]
              \addplot coordinates {
                (0,12.922)
                (1,12.049)
                (2,7.683)
                (3,11.621)
                (4,8.049)
                (5,8.031)
                (6,8.179)
                (7,8.259)
                (8,6.926)
                (9,8.328)
                (10,8.600)
                (11,7.962)
                (12,7.636)
                (13,7.123)
                (14,8.249)
              };
              \end{axis}
            \end{tikzpicture}
         \caption{Worst Negative Slack Time}
         \label{fig:resconsumptionwnst}
     \end{subfigure}
  \caption{Result of the static analysis for different amounts of face of the geo-fence.}
        \label{fig:three graphs}
\end{figure}
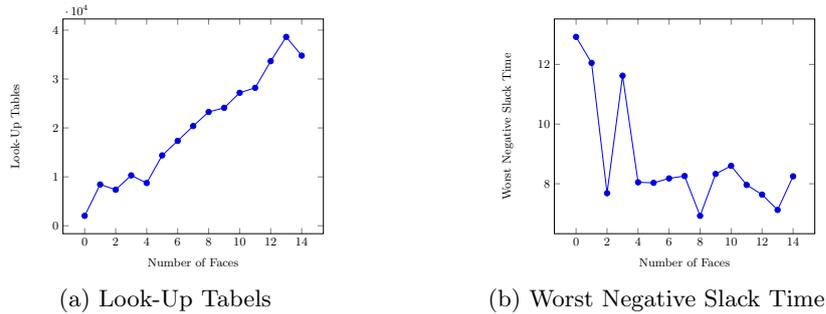

Note that the geo-fence specification checks for 12 intersections in parallel, one for each face of the fence (cf. \cref{fig:mission}). 
Adapting the number of faces allows for scaling the amount of \fpga resources required, as can be seen in \cref{fig:resconsumptionluts}.  
The graph does not grow linearly because the realization problem of \vhdl code onto an \fpga is a multi-dimensional optimization problem with several pareto-optimal solutions.
Under default settings, the optimizer found a solution for four faces that required fewer LUTs than for three faces. 
At the same time, the worst negative slack time (WNST) of the four-face solution was lower than the WNST for the three-face solution as well (cf. \cref{fig:resconsumptionwnst}), indicating that the former performs worst in terms of running time.

\section{Results}\label{sec:eval}

As the title of the paper suggests, the superARTIS with the \rtlola monitor component is cleared to fly and a flight test is already scheduled.
In the meantime, the monitor was validated on log files from past missions of the superARTIS replayed under realistic conditions.
During a flight, the controller polls samples from sensors, estimates the current position, and sends the respective data to the logger and monitor.
In the replay setting, the process remains the same except for one detail: 
Rather than receiving data from the actual sensors, the data sent to the controller is read from a past log file in the same frequency in which they were recorded.
The timing and logging behavior is equivalent to a real execution. 
This especially means that the replayed data points will be recorded again in the same way. 
Control computations take place on a machine identical to the one on the actual aircraft.
As a result, from the point of view of the monitor, the replay mode and the actual flight are indistinguishable. 
Note that the setup is open-loop, \ie, the monitor cannot influence the running system.
Therefore, the replay mode using real data is more realistic than a high-fidelity simulation.

When monitoring the geo-fence of the reconnaissance mission in \cref{fig:mission}, all twelve face crossings were detected successfully.  
Additionally, when replaying the sensor data of the experiment in the enclosed backyard from \cref{sec:paths}, the erratic GPS sensor data lead to 113 violations regarding the GPS module on its own.
Note that many of these violations point to the same culprit:  a low number of available GPS satellites, for example, correlates with the occurrence of peaks in the GPS velocity.
Moreover, the cross validation issued another 36 alarms due to a divergence of IMU and GPS readings. 
Other checks, for example detecting a deterioration of the GPS module based on its output frequency, were not violated in either flight and thus not reported.

\section{Conclusion}
\label{sec:conclusion}
We have presented the integration of a hardware-based monitor into the superARTIS UAV. 
The distinguishing features of our approach are the high level of expressiveness of the \rtlola specification language combined with the formal guarantees on the resource usage. 
The comprehensive tool framework facilitates the development of complex specifications, which can be validated on log data before they get translated into a hardware-based monitor.
The automatic analysis of the specification derives the minimal requirements on the development board needed for safe operation. 
If they are met, the specification is realized on an \fpga and integrated into the superARTIS architecture.
Our experience shows that the overall system works correctly and reliably, even without thorough system-level testing. 
This is due to the non-interfering instrumentation, the validated specification, and the formal guarantees on the absence of dynamic failures of the monitor.
\newpage


\bibliographystyle{splncs04}
\bibliography{bibliography}

\appendix
\newpage
\section{RTLola Specifications}\label{appendix:specs}
This section gives more details on the \rtlola specification. 

\subsection{Sensor Validation}\label{appendix:validation}
In this subsection, we consider GPS receiver submodules, namely: GPS velocity and position.
In the specification shown in \cref{fig:gnssSensorValidation}, the outputs of both modules are declared as inputs to the monitor (Lines 5--6 and 15--18).
Basic checks whether the absolute values of the horizontal and vertical speed are within bounds are performed in lines 9 and 10, where triggers check if any bound is exceeded.
Similarly, valid states of the \lstinline{position_type} (Lines 21) and bounds for \lstinline{diff_age} and \lstinline{solution_age} (lines 23--24) are easily checked.
Lines 26--33 evaluate the performance of the position estimation based on the number of satellites in reach.
In line 27, we count once per second, \ie with $1\hertz$, the number of satellites over the last three seconds.
We expect to see at least nine satellites within this frame of time.
Next, we trigger a notification on each rising edge of \lstinline{behavior_num_sats} (line 32--33).
The stream \lstinline{behavior_num_sats} is an aggregation over five seconds where \lstinline{curr_num_sats_invalid} represents an auxiliary stream which valides the number of satellites.
Once per second we check if within the last five seconds more than twelve events reported a number of satellites below 9.
If this is the case, we warn the system because the GPS modules seems to deteriorate.

The specification in \cref{fig:peakdetection} complements the GPS validation.
Here, we focus on the horizontal (lines 9--11), vertical speed (lines 13--15), and the length of the speed vector (lines 17--19).
Each second, we aggregate the respective speed and take the average.
Each time we receive a new speed event, we compare its value against the computed average value of the last ten seconds.
If there is a strong deviation, we raise a violation.

The validation specification is the combination of \cref{fig:gnssSensorValidation} and \cref{fig:peakdetection}.

\begin{figure}
	\centering
	\small
	\input{specifications/gnssSensorValidation}
	\caption{Sensor validation}
	\label{fig:gnssSensorValidation}
\end{figure}

\begin{figure}
	\centering
	\small
	\input{specifications/peakDetection}
	\caption{Peak detection}
	\label{fig:peakdetection}
\end{figure}

\subsection{Geo-fence}\label{appendix:geofence}
In this section, we depict the detection of a single line crossing of a geo-fence. 
The specification can be seen in \cref{fig:geofence} where italic variables represent constant points of the geo-fence (\textit{lat}, \textit{lon}) and their respective slopes (\textit{m}) and y-intercepts (\textit{b}). 
The basic idea is to compute the intersection of the geo-fence line and the vehicle line, i.e.~the line given the last position and the current position of the vehicle.
The vehicle line is computed in lines 12--21.
The slope of the line and the y-intercept are calculated in lines 20 and 21, respectively.
Given these parameters, the computation of the intersection point is basic geometry, \ie $y = m\cdot x + t$.
Finally, we check whether the intersection point lies on both the vehicle line and the geo-fence face (lines 34--41).
The output \lstinline{is_fnc} is true if the vehicle movement can be encoded as a valid function (line 32).

\begin{figure}
	\centering
	\small
	\input{specifications/geofence}
	\caption{Geo-fencing}
	\label{fig:geofence}
\end{figure}

\subsection{Sensor Cross-Validation}\label{appendix:cross}
In this section, we consider two complementary sensors.
The GPS receiver based on satellite signals and an Inertial Measurement Unit (IMU) based on acceleration and angular rate data.
In the specification given in \cref{fig:crossSpec}, we use the fact that the acceleration is the derivative of the velocity.
This correlation should be present when comparing IMU and GPS receiver events.
Since the IMU frequency is known, we compute the expected velocity, by averaging instead of an integration with the trapezoid abstraction (lines 15--16).
This removes undesired noise.
The average velocity given by the GPS receiver is computed in a straight forward manner (lines 19--22).
A warning is raised if the difference in velocity is greater than 0.5 (line 25). 

\begin{figure}
  \centering
  \small
  \input{specifications/crossValidation}
  \caption{Cross-validation}
  \label{fig:crossSpec}
\end{figure}

\end{document}